\documentclass[sigconf]{acmart}
\usepackage{algorithm}
\usepackage{algorithmic}
\usepackage{tabularx}
\usepackage{colortbl}
\AtBeginDocument{%
  }

\setcopyright{acmlicensed}
\copyrightyear{2026}
\acmYear{2026}
\setcopyright{cc}
\setcctype{by-nc-nd}
\acmConference[WWW '26]{Proceedings of the ACM Web Conference 2026}{April 13--17, 2026}{Dubai, United Arab Emirates}
\acmBooktitle{Proceedings of the ACM Web Conference 2026 (WWW '26), April 13--17, 2026, Dubai, United Arab Emirates}
\acmPrice{}
\acmDOI{10.1145/3774904.3792793}
\acmISBN{979-8-4007-2307-0/2026/04}
\begin{document}


\title{LaV-CoT: Language-Aware Visual CoT with Multi-Aspect Reward Optimization for Multilingual Text-Centric VQA}
\author{Jing Huang}
\email{jh.jj@antgroup.com}
\orcid{0009-0005-5651-9033}
\affiliation{%
  \institution{Ant Digital Technologies, Ant Group}
  \city{Singapore}
  \country{Singapore}
}
\authornote{Both authors contributed equally to the paper.}

\author{Zhiya Tan}
\email{zhiya001@e.ntu.edu.sg}
\orcid{0009-0004-2433-1389}
\affiliation{%
  \institution{College of Computing and Data Science, Nanyang Technological University}
  \city{Singapore}
  \country{Singapore}
}
\affiliation{%
  \institution{Ant Digital Technologies, Ant Group}
  \city{Singapore}
  \country{Singapore}
}
\authornotemark[1]

\author{Shutao Gong}
\email{gongshutao.gst@digital-engine.com}
\orcid{0009-0009-5643-0441}
\affiliation{%
  \institution{Ant Digital Technologies, Ant Group}
  \city{Changsha}
  \country{China}
}

\author{Fanwei Zeng}
\email{fanwei.zfw@antgroup.com}
\orcid{0009-0005-7872-8610}
\affiliation{%
  \institution{Ant Digital Technologies, Ant Group}
  \city{Hangzhou}
  \country{China}
}

\author{Joey Tianyi Zhou}
\email{joey_zhou@cfar.a-star.edu.sg}
\orcid{0000-0002-4675-7055}

\affiliation{%
  \institution{CFAR, Agency for Science, Technology and Research}
  \city{Singapore}
  \country{Singapore}
}

\affiliation{%
  \institution{IHPC, Agency for Science, Technology and Research}
  \city{Singapore}
  \country{Singapore}
}

\author{Changtao Miao}
\email{miaochangtao.mct@antgroup.com}
\orcid{0000-0002-7634-9992}
\affiliation{%
  \institution{Ant Group}
  \city{Hangzhou}
  \country{China}
}
\authornote{Corresponding authors.}

\author{Huazhe Tan}
\email{huazhe.thz@antgroup.com}
\affiliation{%
  \institution{Ant Digital Technologies, Ant Group}
  \city{Beijing}
  \country{China}
}

\author{Weibin Yao}
\email{wenjing.ywb@antgroup.com}
\affiliation{%
  \institution{Ant Digital Technologies, Ant Group}
  \city{Singapore}
  \country{Singapore}
}

\author{Jianshu Li}
\email{jianshu.l@antgroup.com}
\orcid{0000-0001-8554-6886}
\affiliation{%
  \institution{Ant Digital Technologies, Ant Group}
  \city{Singapore}
  \country{Singapore}
}
\authornotemark[2]

\renewcommand{\shortauthors}{Jing Huang et al.}


\begin{abstract}
  Multilingual Text-Centric Visual Question Answering (TEC-VQA) has become crucial for real-world applications, as it requires fine-grained understanding and reasoning over multilingual scene text. Recent advances in vision-language models (VLMs) have demonstrated strong potential in tackling multimodal tasks. However, most existing approaches rely primarily on textual Chain-of-Thought (CoT) and provide limited support for multilingual multimodal reasoning. To address this gap, we introduce \textbf{LaV-CoT}, the first Language-aware Visual CoT framework with Multi-Aspect Reward Optimization. LaV-CoT incorporates an interpretable multi-stage reasoning pipeline consisting of text summary with bounding box, language identification, spatial object-level captioning, and step-by-step logical reasoning. To improve reasoning accuracy and cross-lingual generalization, we propose a novel verifiable Multi-Aspect Reward Optimization in addition to supervised fine-tuning that incorporates rewards for linguistic consistency, structural fidelity, and response accuracy. Extensive evaluations on public datasets, including MMMB, Multilingual MMBench, and MTVQA, show that LaV-CoT outperforms open-source models of similar size by up to \(\sim\)9.5\% accuracy, even surpassing open-source models more than twice its size, and further exceeding several state-of-the-art proprietary models. Moreover, LaV-CoT has been integrated into our online Intelligent Document Processing platform. A further online A/B test demonstrates an \(\sim\)8.7\% improvement in acceptance rate, validating its effectiveness in industrial deployment and commercial applications. Our code is available at this \href{https://github.com/HJNVR/LaV-CoT}{\textit{repository}}.
\end{abstract}

\ccsdesc[500]{Computing methodologies~Machine learning}
\ccsdesc[500]{Computing methodologies~Computer vision}
\ccsdesc[500]{Computing methodologies~Natural language processing}
\keywords{Multilingual Vision-Language Models; Multilingual Text-Centric VQA; Visual Chain-of-Thought Reasoning; Reinforcement Tuning}


\maketitle
\section{Introduction}
Large Vision-Language Models (VLMs) \cite{li2022blip, alayrac2022flamingo, liu2023visual, lu2024deepseek, wang2024qwen2, wang2024cogvlm, li2024monkey, sun2024parrot, bai2025qwen2, zhu2025internvl3, wang2025internvl35advancingopensourcemultimodal, li2025eagle2buildingposttraining, glm2024chatglmfamilylargelanguage} have advanced rapidly in recent years, demonstrating strong performance on multimodal tasks such as image captioning \cite{li2022blip, Wang2023ImageAA, zhou2019unifiedvisionlanguagepretrainingimage}, text–region grounding \cite{zhang2022glipv2unifyinglocalizationvisionlanguage,wan2024contrastiveregionguidanceimproving} and visual question answering \cite{chen2022pali,zhou2019unifiedvisionlanguagepretrainingimage}. Driven by the growing demand from international industrial applications, multilingual Text-Centric Visual Question Answering (TEC-VQA) \cite{tang2024mtvqa,beyondocr+vqa2021,mishraICDAR19} has emerged as a key task to evaluate models’ ability to interpret and reason over multilingual text in images. Recent VLMs have begun to conduct preliminary evaluations on multilingual TEC-VQA benchmarks. 

\begin{figure}[t]
  \centering
  \includegraphics[width=\linewidth]{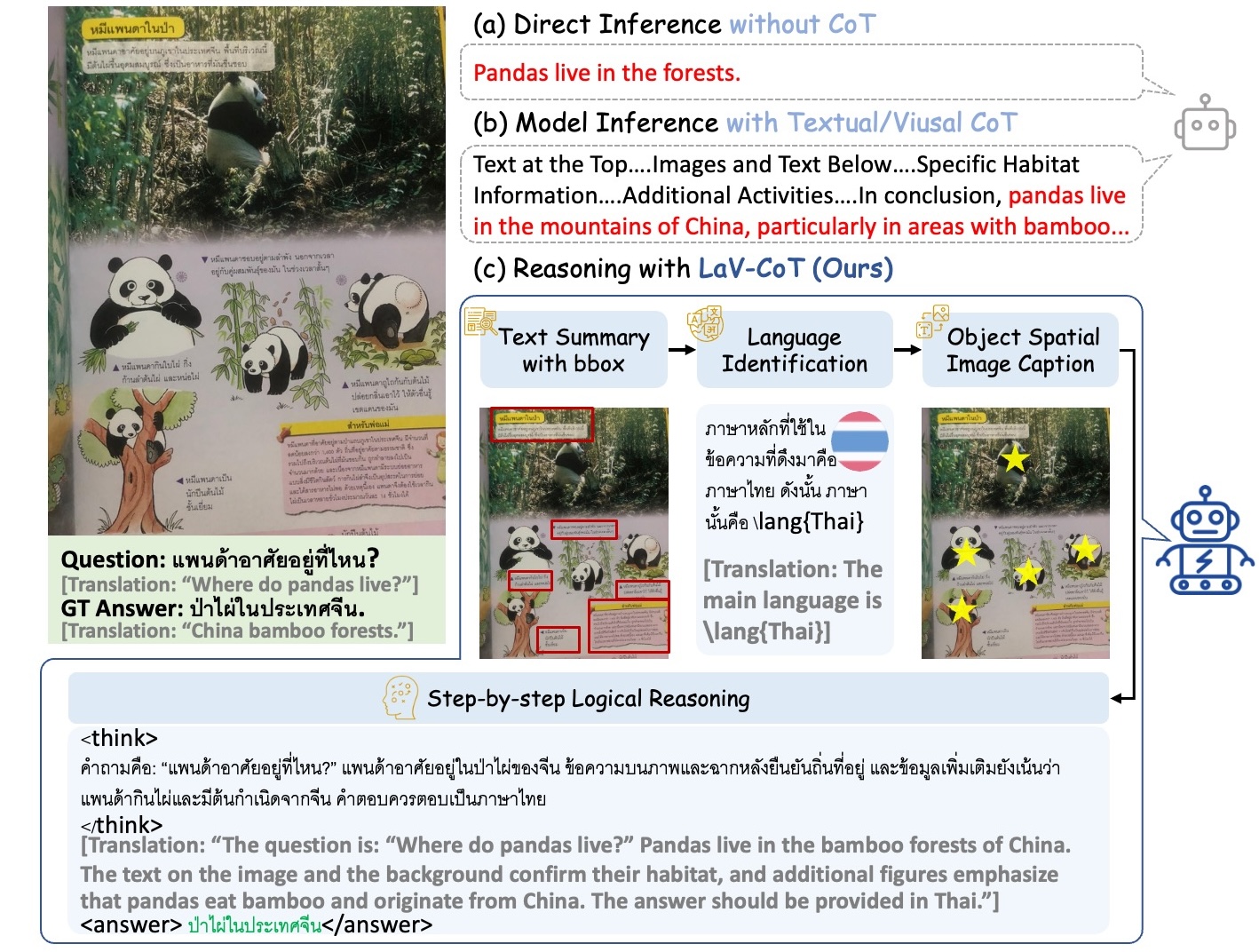}
  \caption{Overview of LaV-CoT: (a) Direct model answers may be incorrect with limited interpretability. (b) Textual or Visual CoT reasoning improves reasoning transparency but still suffers from weak visual grounding and linguistic inconsistency. (c) Our method proposes an interpretable multi-stage reasoning pipeline and gets the correct final answer.}
  \label{fig:overview}
  \Description{Reasoning pipeline}
\end{figure}

Although recent general VLMs are trained on an increasing number of VQA datasets, they still suffer from a scarcity of multilingual text-centric data. Expert models \cite{sun2024parrot, li2024monkey} include additional relevant datasets but remain limited in multimodal reasoning, leaving multilingual TEC-VQA a persistent challenge. As illustrated in Figure~\ref{fig:overview}, three major issues persist:
(i) \textbf{Limited Interpretability}. Direct answers produced by recent VLMs \cite{bai2025qwen2, wang2025internvl35advancingopensourcemultimodal} remain as black boxes, hindering model transparency and interpretability.
(ii) \textbf{Visual–textual misalignment}. Textual CoT approaches \cite{Wei2022ChainOT, xu2025llavacotletvisionlanguage} introduce intermediate natural language reasoning steps to enhance performance. However, these text-only rationales often under-exploit visual cues, leading to weak visual grounding.
(iii) \textbf{Language inconsistency}. Recent Visual CoT methods \cite{shao2024visualcotadvancingmultimodal, zhao2025unsupervisedvisualchainofthoughtreasoning} incorporate visual cues to enable multimodal reasoning, but still suffer from inconsistency across languages.
To address these challenges, we propose LaV-CoT, a \textit{Language-aware Visual CoT reasoning framework with Multi-Aspect Reward Optimization}. LaV-CoT introduces an interpretable multi-stage reasoning pipeline that integrates four complementary components:
(1) \textbf{Text Summary with Bounding Box} \cite{yang2022unitabunifyingtextbox}, which localizes and extracts textual content from the image;
(2) \textbf{Language Identification}, which ensures accurate linguistic understanding and processing;
(3) \textbf{Spatial Object-level Captioning} \cite{li2019visualbertsimpleperformantbaseline}, which grounds visual entities within the scene; and
(4) \textbf{Step-by-step Logical Reasoning} \cite{Suris2023ViperGPT, zhang2024multimodalchainofthoughtreasoninglanguage, xu2025llavacotletvisionlanguage}, where the model systematically analyzes the question and integrates information across modalities to reach the final answer. Together, these components form a structured and interpretable reasoning process, enabling fine-grained cross-modal alignment and improving interpretability in multilingual multimodal tasks. A further challenge lies in constructing high-quality multilingual reasoning data due to the high cost of manual annotation. To this end, we design an automatic data curation method that generates structured and verifiable multilingual multimodal CoT annotations through iterative generation, correction, and refinement \cite{zelikman2022starbootstrappingreasoningreasoning, lightman2023letsverify, shao2024visualcotadvancingmultimodal}, thereby preserving both linguistic fidelity and reasoning quality at scale.

Recent studies have shown that reinforcement learning (RL) leads to better generalization in both rule-based and visual reasoning tasks \cite{chu2025sftmemorizesrlgeneralizes}. However, existing RL approaches are often limited by coarse-grained, binary rewards that fail to capture fine-grained reasoning quality for multilingual TEC-VQA. Therefore, we propose a two-stage paradigm that combines supervised fine-tuning (SFT) \cite{li2022blip, chen2022pali, bai2025qwen2} with language-aware Group Relative Policy Optimization (GRPO) \cite{lu2023deepseekr1, Shao2024DeepSeekMath, liu2025prefix}. Unlike standard optimization methods, our GRPO variant is guided by verifiable multi-aspect rewards, including language consistency, text segment and object count accuracy, final answer edit distance, and format compliance. This multi-aspect reward design enables stable optimization and robust reasoning across diverse languages and modalities. 

Extensive experiments demonstrate that a 3B-parameter model (Qwen2.5-VL-3B) trained under our framework outperforms comparable open-source baselines and advanced proprietary ones in multilingual visual reasoning. A further online A/B testing shows improvements in both answer acceptance rate and user satisfaction.

\noindent \textbf{Our main contributions are as follows:}
\begin{itemize}
    \item We propose LaV-CoT, the first framework to unify language-aware visual CoT reasoning with multi-aspect reward optimization for multilingual text-centric VQA tasks.
    \item We design an automatic data curation method that produces scalable, high-quality multilingual CoT annotations through iterative generation, correction, and refinement. 
    \item We design a language-aware GRPO algorithm with verifiable multi-aspect rewards, enhancing reasoning robustness and cross-modal alignment.
    \item We validate our framework on public benchmarks and through an online A/B test, highlighting both state-of-the-art multilingual multimodal reasoning performance and its effectiveness in real-world deployment.
\end{itemize}

\begin{figure*}[t]  
  \centering
  \includegraphics[width=0.9\textwidth]{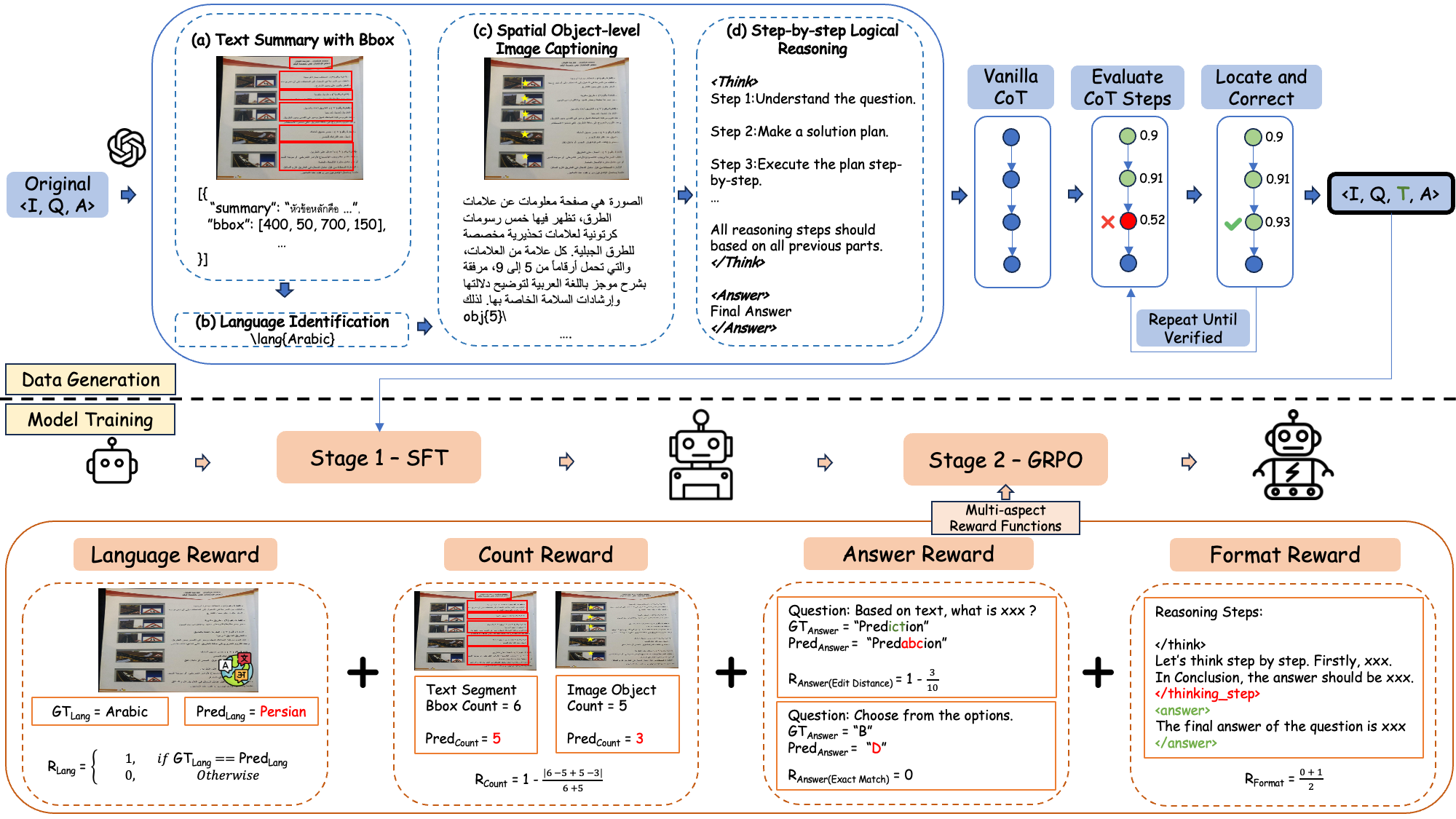}
  \caption{The framework includes an automated data generation pipeline, which leverages a multi-step reasoning process comprising (a) Text Summary with BBox, (b) Language Identification, (c) Spatial Object-level Image Captioning, and (d) Step-by-step Logical Reasoning. This reasoning pipeline first produces vanilla CoT annotations, which are then iteratively refined through rigorous verification to ensure high-quality supervision. The generated data subsequently supports a two-stage training paradigm combining SFT with GRPO. Multi-aspect Reward functions cover language consistency, count accuracy, the edit distance between predicted and ground-truth answers, and format compliance, collectively enhancing structural understanding and reinforcing robust reasoning capabilities.}
  \Description{Illustration of the multi-step data generation pipeline and two-stage training framework integrating SFT and GRPO for multilingual multimodal reasoning.}
  \label{fig:framework}
\end{figure*}

\section{RELATED WORKS}
\subsection{Multilingual Text-Centric VQA in Vision-Language Models}
Recent advances in vision-language models (VLMs) have significantly advanced visual question answering, particularly due to their strong zero-shot generalization capabilities. However, unlike conventional VQA, TEC-VQA represents a more complex challenge, particularly in multilingual settings, where cross-lingual text understanding and visual grounding must be performed jointly. Recent general models, such as Qwen2.5-VL \cite{bai2025qwen2} and InternVL3.5 \cite{wang2025internvl35advancingopensourcemultimodal}, improve multilingual TEC-VQA performance by leveraging publicly available datasets spanning multiple languages. Expert models such as Parrot \cite{sun2024parrot} further target multilingual text-centric reasoning by conditioning visual tokens on diverse language inputs and employing a Mixture-of-Experts (MoE) approach to enhance alignment across multilingual tokens.

\subsection{Chain-of-Thought Reasoning in Vision Language Models}
CoT prompting \cite{Wei2022ChainOT} provides intermediate steps to tackle challenging problems such as commonsense reasoning and logical deduction \cite{cobbe2021trainingverifierssolvemath}. However, textual CoT methods rely solely on linguistic reasoning and often underutilize visual cues, resulting in weak visual grounding. Recent work, such as LLaVA-CoT \cite{xu2025llavacotletvisionlanguage}, employs a structured CoT mechanism that systematically decomposes complex reasoning tasks using textual CoT enhanced with visual inputs, improving multimodal reasoning accuracy. Moreover, studies have shown that integrating CoT strategies significantly enhances the reasoning capability of VLMs, particularly in TEC-VQA. For instance, Visual CoT \cite{shao2024visualcotadvancingmultimodal} improves interpretability by generating bounding boxes that highlight relevant image regions alongside textual explanations, providing clearer visual grounding. Building upon these advancements, we propose a systematic and interpretable language-aware visual CoT pipeline specifically designed to address the challenges of multilingual vision-language reasoning.

\subsection{Visual Reinforcement Fine-Tuning} 
With the emergence of reasoning models such as OpenAI’s o1 \cite{openai2024openaio1card} and DeepSeek-R1 \cite{lu2023deepseekr1}, research on large language models (LLMs) has increasingly focused on enhancing reasoning capabilities through reinforcement learning (RL). Recent studies extend this paradigm to vision-language models (VLMs) via RL-based post-training, known as Visual Reinforcement Fine-Tuning (Visual-RFT). For instance, Visual-RFT~\cite{liu2025visualrftvisualreinforcementfinetuning} extends reinforcement fine-tuning to visual domains by training VLMs with multiple reasoning trajectories and optimizing them using visually verifiable rewards and policy gradient methods such as GRPO. Furthermore, Visionary-R1 \cite{xia2025visionaryr1mitigatingshortcutsvisual} introduces a reinforcement learning framework that encourages visual understanding prior to reasoning by employing a structured caption–reason–answer training paradigm. In our work, we design verifiable multi-aspect reward functions to provide more granular and reliable feedback, enhancing reinforcement learning for multilingual TEC-VQA.

\section{Method}
As illustrated in Figure~\ref{fig:framework}, our framework supports a progressive, step-by-step reasoning process that enhances the reasoning capabilities of VLMs. Building on the reasoning traces, we introduce an automated data curation pipeline that iteratively evaluates and refines each CoT sequence to ensure high-quality training data. Finally, a two-stage training strategy with a novel multi-aspect reward design is applied to produce the fully optimized model.

\begin{table}[tp]
  \centering
  \caption{The number of samples selected from each benchmark, along with the language categories covered.}
  \label{tab:mr-vqa}
  \begin{tabular}{lll}
    \toprule
    \textbf{Dataset}    & \textbf{Language}  & \textbf{Size} \\ 
    \midrule
    COCO2017         & EN,ZH,PT,AR,TR,RU & 73k \\
    Visual-Genome    & EN,ZH,PT,AR,TR,RU & 18k \\
    GQA              & EN,ZH,PT,AR,TR,RU & 15k \\
    OCR-VQA          & EN,ZH,PT,AR,TR,RU & 16k \\
    TextVQA          & EN,ZH,PT,AR,TR,RU & 4.7k \\
    Llava-Pretrain   & EN,ZH,PT,AR,TR,RU & 0.6k \\
    MTVQA            & AR,DE,FR,IT,JA,KO,RU,TH,VI & 21k \\
    \bottomrule
  \end{tabular}
\end{table}

\subsection{Data Generation}
\subsubsection{Multi-stage Reasoning}
 The multi-stage reasoning design is motivated by the way humans naturally approach image understanding: when examining a document image, we first locate salient text regions and retain their summarized content, then identify the language, recognize objects and their spatial relationships, and finally integrate all information to reason step by step toward an answer. Here are the four structured stages:

\begin{itemize}
    \item[(a)] \textbf{Text Summary with Bounding Boxes.} For text-centric images, we first detect text segments within the image. Since extracting the complete OCR content may be costly in terms of token usage, we apply a summarization strategy to generate concise yet informative representations of the detected text. The output is formatted as a list of bounding boxes paired with summarized text, where the length of the list is later used as a reward signal during training.
    
    \item[(b)] \textbf{Language Identification.} Following the saying that the best way to learn a foreign language is to think in that language, for images containing textual content in any language, we identify the primary language based on the summarized text obtained in the previous step. The identified target language is explicitly marked using a \verb|\lang{}| tag, which allows reward calculation to directly compare the predicted and ground-truth languages.
    
    \item[(c)] \textbf{Spatial Object-Level Image Captioning.} To capture comprehensive information from the image, we describe not only the main objects but also their spatial positions, thereby providing a structured understanding of the visual scene. In addition, the model outputs a total object count, explicitly marked using an \verb|\obj{}| tag, which provides a quantitative signal that can be evaluated during reward computation.
    
    \item[(d)] \textbf{Step-by-Step Logical Reasoning.} Leveraging the outputs from all previous steps as evidence, we first understand the given question, then devise a detailed solution plan, and finally execute the plan step-by-step until arriving at the final answer.
\end{itemize}

Once trained, our model can autonomously determine when to initiate each stage, requiring no additional prompts, and complete all stages within a single inference pass. This end-to-end structured reasoning process improves robustness and effectiveness while enabling multi-aspect reward computation, including bounding-box count, \verb|\lang{}| tag correctness, and \verb|\obj{}| count accuracy.

\begin{algorithm}[t]
\caption{Verified CoT Data Generation for Input $\langle I, Q, A \rangle$}
\label{alg:verified-cot}
\begin{algorithmic}[1]
\REQUIRE Image $I$, Question $Q$, Answer $A$, Generator $f_{\text{gen}}$, Evaluator $f_{\text{eval}}$, Threshold $\tau$
\ENSURE Verified CoT data $\langle I, Q, T, A \rangle$
\STATE Initialize $T_{\text{init}} \gets f_{\text{gen}}(\langle I, Q, A \rangle)$
\FOR{$i = 1$ \textbf{to} $|T_{\text{init}}|$}
    \STATE $s_i \gets T_{\text{init}}[i]$
    \STATE $score \gets f_{\text{eval}}(s_i)$
    \WHILE{$score < \tau$}
        \STATE $s_{\text{error}} \gets \text{Locate}(f_{\text{eval}}(s_i))$
        \STATE $s_i \gets \text{Correct}(f_{\text{gen}}(s_{\text{error}}))$
        \STATE Update $T_{\text{init}}[i] \gets s_i$
        \STATE $score \gets f_{\text{eval}}(s_i)$
    \ENDWHILE
\ENDFOR
\STATE \textbf{return} $\langle I, Q, T, A \rangle$
\end{algorithmic}
\end{algorithm}

\subsubsection{Dataset Curation}
Most existing multilingual VQA datasets lack the detailed reasoning processes necessary to effectively train the multilingual reasoning VLM model. To address this limitation, we compile a new dataset by integrating samples from several widely used VQA benchmarks, resulting in a total of 148k image–question–CoT–answer pairs, as illustrated in Table~\ref{tab:mr-vqa}. 

Specifically, as shown in Algorithm~\ref{alg:verified-cot}, we start from the original question--answer triplets $\langle I, Q, A \rangle$, where $I$ denotes the image, $Q$ the question, and $A$ the corresponding answer. We first prompt a generator $f_{\text{gen}}$ to produce an initial sequence of vanilla CoT steps $T_{\text{init}} = \{s_i\}_{i=1}^{|T_{\text{init}}|}$. Next, we prompt a evaluator $f_{\text{eval}}$ to score each step $s_i \in T_{\text{init}}$. Both $f_{\text{gen}}$ and $f_{\text{eval}}$ use \textit{GPT-4o-2024-05-13} \cite{hurst2024gpt} with default decoding settings. For any step whose score falls below the threshold $\tau$, we iteratively 
perform the following procedure: First, we apply the evaluator to the step and then locate the erroneous part, denoted as $s_{\text{error}}$, using the function $\text{Locate}(f_{\text{eval}}(s_i))$. Next, a corrected step $s_i$ is generated by applying the function $\text{Correct}$ to $s_{\text{error}}$, i.e., $s_i$ is updated as $\text{Correct}(f_{\text{gen}}(s_{\text{error}}))$. The corrected step then replaces the original step in the sequence $T_{\text{init}}$. Finally, the updated step is re-evaluated to obtain the new score using $f_{\text{eval}}(s_i)$.
This evaluation--correction--update loop repeats until all steps in $T_{\text{init}}$ exceed the threshold. The final verified Chain-of-Thought sequence is denoted as $T$, and the output dataset consists of $\langle I, Q, T, A \rangle$. Detailed prompt designs
are shown in Appendix~\ref{app:prompt_design}.

Our dataset covers 13 languages, including English (EN), Chinese (ZH), Portuguese (PT), Arabic (AR), 
Turkish (TR), Russian (RU), German (DE), French (FR), Italian (IT), Japanese (JA), Korean (KO), Thai (TH), 
and Vietnamese (VI). These languages represent a diverse range of linguistic families. Our dataset is constructed from diverse sources, including COCO2017 \cite{Lin2014MicrosoftCO}, a large-scale dataset for object detection, segmentation, and image captioning containing everyday scenes with common objects in context; Visual Genome \cite{krishna2016visual}, which provides densely annotated images linking objects, attributes, relationships, and region descriptions for visual reasoning; GQA \cite{hudson2019gqa}, emphasizing compositional reasoning over real-world images with complex question-answer pairs; OCR-VQA \cite{mishraICDAR19} and TextVQA \cite{singh2019towards}, both focusing on reading and reasoning over text within images for text-centric visual question answering; Llava-Pretrain \cite{lin2023llava}, a multimodal pretraining dataset supporting various VQA and reasoning tasks across multiple domains; and MTVQA \cite{tang2024mtvqa}, a multilingual text-based VQA dataset designed to evaluate cross-lingual multimodal understanding.

\subsection{Model Training}
\subsubsection{Two-stage Training
Paradigm}
In the first stage, we adopt SFT to equip the model with basic multilingual and multimodal reasoning abilities, followed by GRPO-based RL post-training in the second stage to further enhance its performance.
However, standard GRPO rewards are typically coarse-grained and binary (e.g., assigning 1 for a correct answer and 0 otherwise), failing to capture the fine-grained reasoning quality required for multilingual TEC-VQA.
Moreover, conventional GRPO provides no explicit supervision over the reasoning process itself, such as maintaining language consistency or ensuring coherent step-by-step reasoning quality. To further refine output quality, we introduce a composite reward function, $R_{\mathrm{Multi\_Aspect}}$, which aggregates multiple rule-based criteria into a single supervisory signal.
\subsubsection{Multi-aspect Reward}
The multi-aspect reward $R_{\mathrm{Multi\_Aspect}}$ consists of four complementary components: three novel rewards designed to capture different aspects of multilingual visual-textual reasoning, and a format reward that ensures the generated output adheres to the expected structural conventions.

\textbf{1. Language Consistency Reward ($R_{\mathrm{Lang}}$).} To encourage the model to perform reasoning in the target language, we compare the language predicted by the model with the labeled primary language of the input. Let $L$ denote the ground-truth language label and $\hat{L}$ represent the language identified by the model. The reward is defined as:
\[
R_{\mathrm{Lang}} =
\begin{cases}
1, & \text{if } L = \hat{L} \\
0, & \text{otherwise}
\end{cases}
\]

\textbf{2. Text Segments and Object Count Reward ($R_{\mathrm{Count}}$).} To ensure accurate text segmentation and object count, let $N_{Ts}$ and $\hat{N}_{Ts}$ be the numbers of reference and predicted text segments, and $N_{Obj}$ and $\hat{N}_{Obj}$ be the numbers of reference and predicted main objects. The reward is defined as:
{\small
\[
R_{\mathrm{Count}} =
\begin{cases}
1 - \dfrac{\left| (N_{Ts} - \hat{N}_{Ts}) + (N_{Obj} - \hat{N}_{Obj}) \right|}{N_{Ts} + N_{Obj}}, 
& \text{if } N_{Ts} + N_{Obj} > 0, \\
1, 
& \text{otherwise}
\end{cases}
\]
}

\textbf{3. Edit Distance of Final Answer Reward ($R_{\mathrm{Answer}}$).}  
We compute the normalized Levenshtein distance $D(Y, \hat{Y})$ between the reference answer $Y$ and the model prediction $\hat{Y}$:
\[
R_{\mathrm{Answer}} = 1 - \frac{D(Y, \hat{Y})}{\max(\ell_Y, \ell_{\hat{Y}})}
\]


\textbf{4. Format Reward ($R_{\mathrm{Format}}$).} 
This reward encourages the model to generate outputs that adhere to the prescribed structural format. 
Let $\hat{Y}$ denote the model output, and let $T = \{(t_i, w_i)\}$ be the set of required tag pairs $t_i$ with weights $w_i \in [0,1]$ such that $\sum_i w_i = 1$, e.g., \texttt{<think></think>} and \texttt{<answer></answer>}. The format reward is computed as a weighted sum over all tags:
\[
R_{\mathrm{Format}} = \sum_{(\scriptstyle t_i, w_i) \in T} w_i \cdot \mathbf{1}_{t_i}(\hat{Y})
\], where the indicator function is defined as: \[
\mathbf{1}_{t_i}(\hat{Y}) =
\begin{cases}
1, & \text{if the tag pair } t_i \text{ appears completely in } \hat{Y},\\
0, & \text{otherwise.}
\end{cases}
\]


The final multi-aspect reward is a weighted combination of these four components:
\[
R_{\mathrm{Multi\_Aspect}} = \alpha \, R_{\mathrm{Lang}} + \beta \, R_{\mathrm{Count}} + \gamma \, R_{\mathrm{Answer}} + \delta \, R_{\mathrm{Format}},
\]
where $\alpha$, $\beta$, $\gamma$, and $\delta$ are non-negative coefficients that control the relative importance of each reward component.

The objective function of GRPO \cite{lu2023deepseekr1} is defined as:
\begin{align}
\mathcal{J}_{\mathrm{GRPO}}(\theta) 
&= \frac{1}{G} \sum_{i=1}^G \Bigg[
\min \Big(
\frac{\pi_\theta(o_i \mid q)}{\pi_{\theta_{\mathrm{old}}}(o_i \mid q)} A_i, \notag\\
&\qquad\quad
\mathrm{clip}\Big(\frac{\pi_\theta(o_i \mid q)}{\pi_{\theta_{\mathrm{old}}}(o_i \mid q)}, 1-\epsilon, 1+\epsilon\Big) A_i
\Big) \Bigg] \notag\\
&\quad - \beta \, \mathbb{D}_{\mathrm{KL}}(\pi_\theta \,\|\, \pi_{\mathrm{ref}})
\end{align}
where now the advantage $A_i$ is computed as:
\[
A_i = \frac{R_{\mathrm{Multi\_Aspect}}(o_i) - \mathrm{mean}\big(\{R_{\mathrm{Multi\_Aspect}}(o_j)\}_{j=1}^G\big)}
{\mathrm{std}\big(\{R_{\mathrm{Multi\_Aspect}}(o_j)\}_{j=1}^G\big) + \varepsilon}
\]

The GRPO objective evaluates how well the current policy generates outputs compared to the previous policy. Probabilities under the current and previous policies are compared for importance sampling, while a clipping mechanism and KL-divergence penalty ensure stable updates and prevent excessive deviation from a reference policy. 
By integrating multi-aspect rewards covering language consistency, segmentation, object count, correctness, and format, our refined GRPO provides fine-grained supervision for multilingual multimodal reasoning.

\begin{figure*}[ht]  
  \centering
  \includegraphics[width=0.9\textwidth]{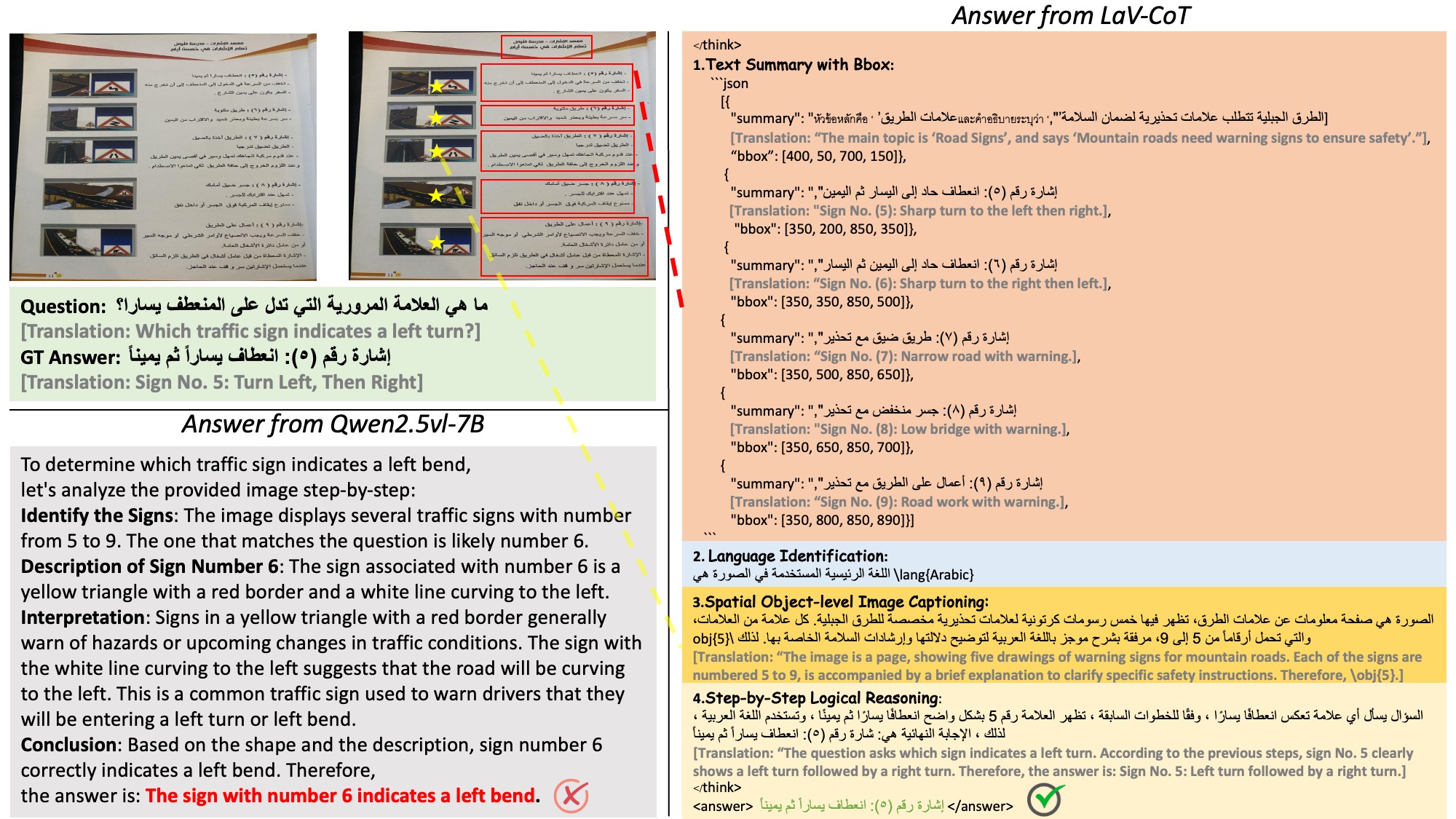}
  \caption{Comparison between Qwen2.5-VL-7B and LaV-CoT. As illustrated, Qwen2.5-VL-7B demonstrates a step-by-step reasoning process; however, it fails to perform the reasoning in the target Arabic language and produces an incorrect final deduction. In contrast, LaV-CoT effectively follows its reasoning pipeline to produce the correct final answer.}
  \label{fig:demo_case}
  \Description{demo case}
\end{figure*}

\section{Experiment}
In this section, we provide an overview of the experimental framework, including implementation details, evaluation benchmarks, and baseline models used for comparison. 
\subsection{Experiment Setting}
\subsubsection{Implementation Details.}
During the Supervised Fine-Tuning (SFT) stage, the model was trained for 1–3 epochs with a global batch size of 256 on 8 NVIDIA A100 (40GB) GPUs. The image resolution size is set to be 896 * 896 \cite{li2024monkey}. The learning rate was fixed at $2 \times 10^{-4}$ and kept constant without decay, using the fused AdamW optimizer. Mixed-precision training with bfloat16 (bf16) and TensorFloat-32 (tf32) was enabled to improve computational efficiency while maintaining numerical stability. Gradient clipping with a maximum norm of 0.3 and a warm-up ratio of 0.03 were applied. Parameter-efficient fine-tuning was implemented via LoRA\cite{hu2021lora}, targeting the transformer modules \texttt{q\_proj}, \texttt{k\_proj}, \texttt{v\_proj}, \texttt{o\_proj}, \texttt{gate\_proj}, \texttt{up\_proj}, and \texttt{down\_proj}, with rank 32, alpha 64, and a dropout rate of 0.05. In the subsequent GRPO training stage, all reward components were uniformly scaled by 0.25 so that the total reward summed to 1, i.e.,
\[
R_{\mathrm{Multi\_Aspect}} = 0.25\,R_{\mathrm{Lang}} + 0.25\,R_{\mathrm{Count}} + 0.25\,R_{\mathrm{Answer}} + 0.25\,R_{\mathrm{Format}}
\]
The number of generations per iteration was set to 4, while other hyperparameters were kept consistent with the SFT stage.

\subsubsection{Evaluation Benchmark.}
For our experiments, MMMB \cite{sun2024parrot}, Multilingual MMBench \cite{sun2024parrot}, and MTVQA \cite{tang2024mtvqa} serve as the primary evaluation benchmarks. MTVQA is designed to specifically measure a model’s ability to align textual and visual information in multilingual text-centric VQA scenarios. To verify LaV-CoT's broader applicability across multilingual VQA scenarios, we additionally evaluate on MMMB and Multilingual MMBench, which target general multilingual multimodal reasoning and a wide variety of question types. Together, these datasets form a complementary evaluation suite that assesses both broad multilingual reasoning capabilities and multilingual multimodal text understanding. For evaluation, we adopt VLMEvalKit \cite{duan2024vlmevalkit} with consistent configuration settings, thereby ensuring a fair and consistent comparison.

\subsubsection{Comparison Models.}
\begin{sloppypar}
We adopt Qwen2.5-VL-3B as our backbone model and train it under the LaV-CoT framework. For a comprehensive evaluation, we compare it against a diverse set of methods, categorized by model type. General models include open-source models of comparable size, such as Qwen2-VL-2B \cite{wang2024qwen2}, InternVL3-2B \cite{zhu2025internvl3}, InternVL3.5-2B \cite{wang2025internvl35advancingopensourcemultimodal} and Qwen2.5-VL-3B \cite{bai2025qwen2}, as well as larger open-source models with more than twice the number of parameters, including DeepSeek-VL-7B \cite{lu2024deepseek}, GLM-4v-9B \cite{glm2024chatglmfamilylargelanguage}, Qwen2-VL-7B \cite{wang2024qwen2}, Qwen2.5-VL-7B \cite{bai2025qwen2}, InternVL3-8B \cite{zhu2025internvl3} and InternVL3.5-8B \cite{wang2025internvl35advancingopensourcemultimodal}. We also include much larger proprietary frontier models, such as GPT-4o-2024-05-13 \cite{hurst2024gpt} and Gemini-2.5-flash \cite{gemini2023}. In addition, we compare with expert models that are either specialized for text-rich visual reasoning and multilingual VQA, which include BlueLM-V-3B \cite{xiong2025bluelm253btechnicalreport} Monkey-9.8B \cite{li2024monkey}, LLaVA-OneVision-7B \cite{Li2024LLaVAOneVision}, PARROT-7B \cite{sun2024parrot}, and Eagle2-9B \cite{li2025eagle2buildingposttraining}.
\end{sloppypar}
\setlength{\tabcolsep}{2pt} 
\renewcommand{\arraystretch}{1} 
\definecolor{lightgray}{gray}{0.9}
\begin{table*}[t]
    \centering
    \scriptsize
    \begin{tabularx}{\textwidth}{l|*{6}{>{\centering\arraybackslash}X}|*{6}{>{\centering\arraybackslash}X}|*{9}{>{\centering\arraybackslash}X}|>{\centering\arraybackslash}X}
        \toprule
        \textbf{} 
        & \multicolumn{6}{c|}{\textbf{MMMB}} 
        & \multicolumn{6}{c|}{\textbf{Multilingual MMBench}} 
        & \multicolumn{9}{c|}{\textbf{MTVQA}} 
        & \\ 
        \rowcolor{lightgray}
        {\tiny \cellcolor{lightgray}\textbf{Model Name}} 
        & en & zh & pt & ar & tr & ru
        & en & zh & pt & ar & tr & ru
        & ar & de & fr & it & ja & ko & ru & th & vi
        & {\tiny \cellcolor{lightgray}\textbf{Overall}} \\ 
        \midrule
        \multicolumn{22}{c}{\textit{General Models}} \\ 
        \midrule
        Qwen2-VL-2B & 78.3 & 74.2 & 72.6 & 68.3 & 61.8 & 72.8 & 72.1 & 71.1 & 69.9 & 61.1 & 54.4 & 69.3 & 7.1 & 27.0 & 27.5 & 32.6 & 12.9 & 23.7 & 11.0 & 3.9 & 24.1 & 52.8\\
        Qwen2.5-VL-3B & 81.2 & 81.0 & 74.8 & 71.0 & 65.6 & 75.2 & 81.1 & 79.7 & 75.8 & 69.3 & 62.4 & 72.6 & 11.7 & 27.3 & 33.3 & 31.7 & 12.7 & 26.2 & 10.3 & 10.4 & 35.6 & 57.3 \\
        InternVL3-2B & 81.9 & 78.3 & 75.4 & 68.6 & 62.9 & 74.6 & 81.3 & 77.8 & 75.9 & 66.4 & 59.5 & 70.7 & 5.5 & 34.0 & 41.1 & 39.7 & 19.6 & 31.5 & 9.4 & 16.5 & 28.5 & 57.4\\
        InternVL3.5-2B & 80.2 & 77.7 & 75.9 & 68.5 & 69.1 & 76.3 & 79.0 & 76.5 & 74.3 & 64.4 & 63.1 & 72.2 & 9.2 & 34.5 & 42.3 & 39.8 & 20.9 & 32.7 & 10.8 & 17.9 & 29.8 & 58.0\\
        \hline
        DeepSeek-VL-7B & 72.6 & 65.9 & 64.4 & 49.7 & 49.0 & 67.5 & 70.7 & 64.0 & 62.6 & 48.0 & 47.9 & 65.5 & 1.4 & 18.4 & 20.4 & 18.0 & 5.1 & 7.0 & 1.6 & 3.5 & 7.2 & 43.9\\
        GLM-4v-9B & 69.2 & 62.8 & 61.5 & 47.2 & 46.9 & 64.3 & 67.9 & 61.3 & 60.0 & 46.1 & 45.7 & 63.0 & 7.0 & 31.4 & 39.3 & 37.9 & 11.1 & 13.4 & 8.1 & 8.2 & 26.6 & 46.2\\
        Qwen2-VL-7B & 83.9 & 82.4 & 81.2 & 79.0 & 74.7 & 82.4 & 81.8 & 81.6 & 79.1 & 75.6 & 74.5 & 79.3 & 16.2 & 30.5 & 35.8 & 36.9 & 16.8 & 30.1 & 11.1 & 11.3 & 28.7 & 61.6 \\
        Qwen2.5-VL-7B & 85.0 & 83.6 & 82.1 & 83.3 & 76.4 & 83.2 & 85.3 & 85.8 & 83.0 & 80.2 & 75.7 & 82.9 & 17.8 & 30.5 & 33.3 & 37.2 & 18.5 & 38 & 13.6 & 15.2 & \underline{\textbf{44.8}} & 64.4 \\
        InternVL3-8B & 85.1 & 83.1 & 82.5 & 81.6 & 76.2 & 83.4 & 85.5 & 85.6 & 83.2 & 79.2 & 75.9 & 82.6 & 9.8 & 37.4 & \underline{\textbf{45.5}} & 41.4 & 22.3 & 31.9 & 11.4 & 18.2 & 38.5 & 64.7\\
        InternVL3.5-8B & 84.9 & 83.0 & 81.4 & 79.6 & 77.4 & 83.1 & 84.5 & 85.7 & 80.9 & 82.8 & 75.8 & 82.3 & 16.1 & 38.2 & 44.1 & 40.6 & 22.8 & 33.7 & 11.6 & 20.2 & 39.8 & 64.9\\
        \hline
        GPT-4o-2024-05-13 & 84.9 & 84.3 & 82.8 & 82.3 & 79.0 & 83.3 & 87.6 & 88.2 & 85.5 & 85.6 & 82.9 & 86.2 & 21.3 & 35.1 & 42.2 & 37.2 & 19.9 & 35.1 & \underline{\textbf{15.9}} & 26.0 & 39.6 & 66.6 \\
        Gemini-2.5-flash & 85.2 & 84.5 & \underline{\textbf{83.1}} & 81.8 & 79.6 & \underline{\textbf{84.0}} & 88.3 & 89.0 & \underline{\textbf{86.1}} & 84.2 & 85.8 & \underline{\textbf{88.1}} & 21.6 & 36.8 & 41.7 & \underline{\textbf{43.0}} & 19.8 & 36.3 & 15.8 & \underline{\textbf{26.9}} & 41.7 & 67.4 \\
        \midrule
        \multicolumn{22}{c}{\textit{Expert Models}} \\ 
        \midrule
        BlueLM-V-3B & -- & -- & -- & -- & -- & -- & -- & -- & -- & -- & -- & -- & 17.3 & \underline{\textbf{39.5}} & 44.7 & 32.2 & \underline{\textbf{23.5}} & 34.0 & 9.2 & 20.3 & 22.9 & -- \\
        Monkey-9.8B & 66.0 & 58.1 & 46.3 & 38.8 & 37.6 & 48.5 & 58.0 & 53.5 & 49.5 & 31.0 & 31.3 & 45.1 & 1.1 & 16.4 & 21.2 & 21.7 & 4.3 & 5.4 & 5.3 & 6.1 & 8.4 & 35.0\\
        LLaVA-OneVision-7B & 79.0 & 78.2 & 75.9 & 73.3 & 67.7 & 76.3 & 76.7 & 75.3 & 73.4 & 70.4 & 64.8 & 73.1 & 5.0 & 21.0 & 22.3 & 26.1 & 6.2 & 7.3 & 6.0 & 3.0 & 13.6 & 53.7\\
        PARROT-7B & 80.1 & 80.0 & 79.6 & 76.5 & 75.0 & 79.9 & 78.0 & 77.1 & 76.7 & 75.9 & 74.0 & 77.7 & 14.3 & 30.1 & 34.6 & 36.7 & 16.3 & 20.2 & 9.5 & 9.9 & 27.8 & 59.7\\
        Eagle2-9B & 84.4 & 83.0 & 81.8 & 81.2 & 75.8 & 83.0 & 83.8 & 83.7 & 81.5 & 78.0 & 75.1 & 81.1 & 17.0 & 29.1 & 34.8 & 37.0 & 17.6 & 33.3 & 10.2 & 13.4 & 32.2 & 62.8 \\
        \midrule
        \multicolumn{22}{c}{\textit{Ours}} \\
        \midrule
        \textbf{LaV-CoT-3B (SFT)} & 83.2 & 81.1 & 80.6 & 80.5 & 77.3 & 82.2 & 85.7 & 84.4 & 84.3 & 83.9 & 85.2 & 83.1 & 19.8 & 33.7 & 38.1 & 33.6 & 20.7 & 37.1 & 10.4 & 22.1 & 37.7 & 64.7\\
        \textbf{LaV-CoT-3B (SFT + GRPO)} & \underline{\textbf{86.0}} & \underline{\textbf{84.8}} & 83.0 & \underline{\textbf{82.7}} & \underline{\textbf{80.3}} & 83.6 & \underline{\textbf{89.0}} & \underline{\textbf{89.4}} & 85.9 & \underline{\textbf{88.3}} & \underline{\textbf{86.8}} & 87.7 & \underline{\textbf{23.2}} & 35.3 & 42.8 & 35.9 & 23.1 & \underline{\textbf{38.2}} & 11.5 & 23.8 & 38.8 & \underline{\textbf{67.5}}\\
        \bottomrule
    \end{tabularx}
    \caption{Performance of multilingual vision-language models on MMMB, Multilingual MMBench, and MTVQA across multiple languages. For each language and the overall average accuracy, the best results are highlighted in bold and underlined.}
    \label{tab:results}
\end{table*}
\section{Main Results}
\subsection{Quantitative Analysis}

Table~\ref{tab:results} presents the multilingual evaluation results of various methods based on general and expert vision-language models across three benchmarks: MMMB, Multilingual MMBench, and MTVQA. Several key observations can be made from both overall comparisons and benchmark-specific analyses.
\sloppy
\subsubsection{Overall and comparative performance.} Large proprietary models such as GPT-4o and Gemini-2.5-flash achieve strong performance across most benchmarks, with overall accuracy of 66.6\% and 67.4\%, respectively, establishing a strong upper bound for multilingual multimodal reasoning. Among open-source models,  Qwen2.5-VL-7B, InternVL3.5-8B, and our proposed LaV-CoT variants deliver competitive performance, achieving overall scores of 64.4\%, 64.9\% and 67.5\%, respectively. Notably, LaV-CoT (SFT + GRPO) outperforms all baselines on average across MMMB, Multilingual MMBench, and MTVQA.

Medium-sized models such as Monkey-9.8B, DeepSeek-VL-7B, and GLM-4v-9B remain below 50\% overall, highlighting the challenges of multilingual multimodal reasoning. Smaller-scale models like Qwen2.5-VL-3B, InternVL3-2B and InternVL3.5-2B achieve modest improvements above 50\%, whereas larger open-source systems such as PARROT-7B, LLaVA-OneVision-7B, Qwen2-VL-7B and Eagle2-9B deliver substantially stronger results. Our LaV-CoT (SFT) model outperforms strong baselines, including Qwen2.5-VL-7B and InternVL3.5-8B, and with further reinforcement tuning (SFT + GRPO), it achieves state-of-the-art performance, approaching and slightly exceeding proprietary models.
\begin{figure}[t]
  \centering
  \includegraphics[width=\linewidth]{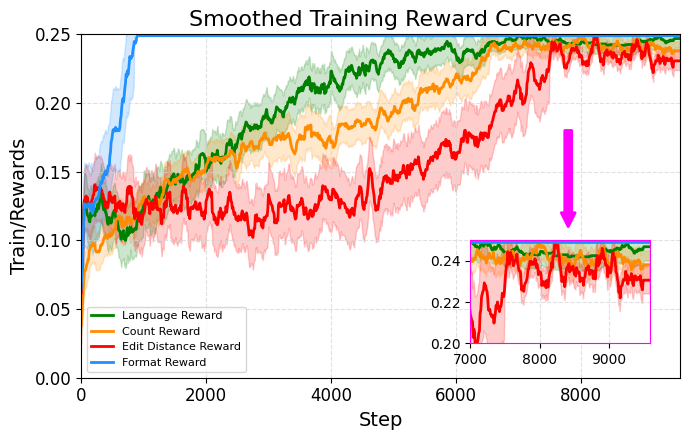}
  \caption{LaV-CoT GRPO Training Reward Curves.}
  \label{fig:rewards}
  \Description{rewards}
\end{figure}


\subsubsection{Benchmark-specific insights.} On MMMB and Multilingual MMBench, LaV‑CoT achieves robust improvements across most languages, with slightly lower scores on Portuguese and Russian.
On MTVQA, our model also delivers substantial gains in low-resource languages such as Arabic and Korean.

Overall, a 3B model trained under the LaV-CoT framework, with multi-aspect reward design and multi-stage reasoning, achieves state-of-the-art performance among open-source multilingual multimodal models and surpasses several proprietary models. We further evaluate additional standard document understanding benchmarks, including DocVQA \cite{mathew2021docvqadatasetvqadocument}, ChartQA \cite{masry2022chartqabenchmarkquestionanswering}, and OCRBenchv2 \cite{fu2025ocrbenchv2improvedbenchmark}, with detailed results in Appendix~\ref{app:add_performance}.


\subsection{GRPO Training Reward Curves Analysis}

Figure \ref{fig:rewards} illustrates the smoothed reward curves during GRPO training, highlighting the evolution of four key reward components: Language Reward, Count Reward, Edit Distance Reward, and Format Reward. The Format Reward exhibits rapid initial improvement, ascending from approximately 0.125 to 0.25 within the first 850 steps before stabilizing, indicating the base model has decent instruction following capability and shows early convergence in format adherence. In contrast, the Language Reward experiences a brief decline from 0.13 to 0.11 by step 800, followed by a steady ascent to 0.25 around step 6500, with subsequent fluctuations reflecting ongoing refinement in linguistic quality. The Count Reward demonstrates a gradual rise from 0.06, including a plateau between steps 3000 and 4000, reaching stability at 0.24 post-step 6540, suggesting progressive optimization of quantitative accuracy. The Edit Distance Reward initially decreases from 0.13 to 0.12 by step 2600, then recovers to 0.22 by step 7500 and converges around 0.23, underscoring challenges in minimizing textual discrepancies early in training. Curves are smoothed using a moving average with a window size of 10, and shaded areas represent ±1 standard deviation, reflecting the variability in reward estimates. Overall, these dynamics demonstrate GRPO's effectiveness in balancing multi-objective rewards, achieving high-fidelity convergence by the end of training.


\section{Ablation Study}
\subsection{Impact of Language-Aware Visual CoT on Performance}
Table \ref{table:lavcot_sft} presents an ablation study comparing different training strategies with Qwen2.5-VL-3B as the baseline. "Direct Train" denotes standard supervised fine-tuning on task-specific data without any intermediate reasoning supervision, and "SFT w/ Text CoT" incorporates textual Chain-of-Thought reasoning. "SFT w/ LaV-CoT" further integrates language-aware visual CoT, resulting in consistent improvements across representative tasks. On average, fine-tuning with Language-Aware Visual CoT achieves a relative improvement of \(\sim\)6.6\% over the baseline across all tasks, highlighting the effectiveness of structured multilingual visual reasoning.

\begin{table}[!t]
\centering
\small
\caption{Ablation study of different fine-tuning methods across various tasks on Multilingual MMBench. Here \textit{OCR} represents Optical Character Recognition, \textit{SR} represents Spatial Relationship, \textit{IR} represents Identity Reasoning, \textit{OL} represents Object Localization, and \textit{SIU} represents Structured Image-Text Understanding.}
\setlength{\tabcolsep}{3pt} 
\begin{tabularx}{\linewidth}{l|XXXXX|X}  
\toprule
{\textbf{Model}} & {\textbf{OCR}} & {\textbf{SR}} & {\textbf{IR}} & {\textbf{OL}} & {\textbf{SIU}} & {\textbf{Avg}} \\
\midrule
Qwen2.5-VL-3B (Baseline) & 83.7 & 68.5 & 74.4 & 72.2 & 75.0 & 74.8 \\
Qwen2.5-VL-3B (Direct Train) & 85.7 & 71.5 & 76.4 & 74.4 & 77.0 & 77.0 \\
Qwen2.5-VL-3B (SFT w/ Text CoT) & 87.7 & 73.5 & 78.4 & 76.1 & 79.0 & 78.9 \\
Qwen2.5-VL-3B (\textbf{SFT w/ LaV-CoT}) & \textbf{88.6} & \textbf{75.2} & \textbf{80.5} & \textbf{81.0} & \textbf{81.5} & \textbf{81.4} \\
\bottomrule
\end{tabularx}
\label{table:lavcot_sft}
\end{table}
\subsection{Comparison of Multi-Aspect Reward Configurations}
\begin{figure}[t]
  \centering
  \includegraphics[width=0.85\linewidth]{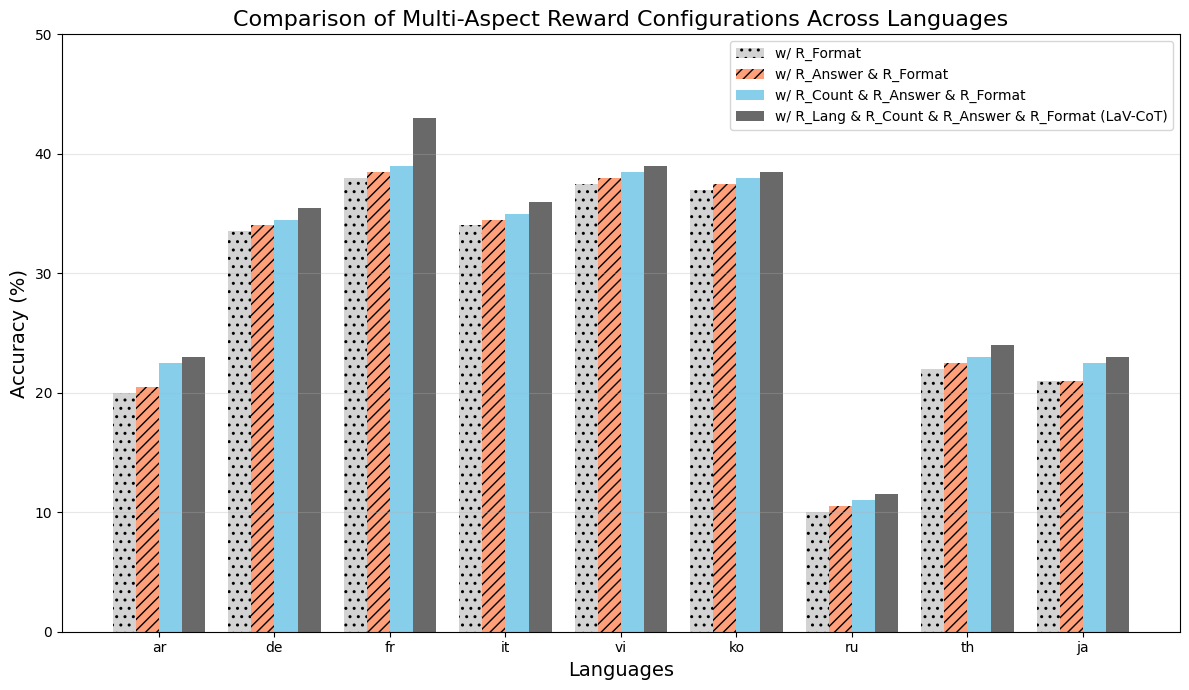}
  \caption{Ablation study of multi-aspect reward configurations on MTVQA. Here \textit{R\_Lang} represents Language Consistency Reward, \textit{R\_Count} represents Text Segments and Object Count Reward, \textit{R\_Answer} represents Edit Distance of Final Answer Reward, and \textit{R\_Format} represents  Format Reward.}
  \label{fig:ab_rewards}
  \Description{ablation rewards}
\end{figure}

To investigate the impact of different reward configurations on model performance, we conduct an ablation study by varying the reward ratios $(\alpha, \beta, \gamma, \delta)$ in the GRPO framework. Figure~\ref{fig:ab_rewards} illustrates the performance across nine representative languages on MTVQA. Specifically, we compare four configurations: training with Format Reward $(0,0,0,1)$, which incorporates only the format reward and achieves accuracy of 28.9\%; training with Answer \& Format Rewards $(0,0,0.5,0.5)$, which covers both answer and format supervision and yields a moderate gain with an average accuracy of 29.36\%; training with Count \& Answer \& Format Rewards $(0,0.33,0.33,0.33)$, which adopts a more balanced setting across count, answer, and format rewards, further improving performance to 30.11\%; and training with all four Rewards (LaV-CoT) $(0.25,0.25,0.25,0.25)$, which jointly optimizes all reward dimensions and delivers the best overall results with an average accuracy of 31.13\%. These findings highlight that balancing multi-aspect rewards is crucial for robust multilingual multimodal reasoning.


\section{Online A/B Test}

We conducted an online A/B test comparing LaV-CoT with the existing production pipeline on our Intelligent Document Processing platform \footnote{https://www.zoloz.com/realdoc} for two weeks. Key business and user-centric metrics were tracked, including answer acceptance rate and user satisfaction scores collected through post-interaction feedback. The results demonstrate that LaV-CoT substantially outperforms the baseline: the answer acceptance rate increased by 8.7\%, and the user satisfaction score improved by 12.4\%, confirming that structured multi-stage reasoning enhances both accuracy and overall user experience in practical document understanding scenarios.
\section{Conclusion}

In this paper, we present LaV-CoT, a novel framework for multilingual Text-Centric VQA. By combining language-aware visual Chain-of-Thought reasoning with structured supervision and GRPO-based multi-aspect rewards, LaV-CoT improves interpretability, visual-text alignment, and language consistency. Extensive experiments across diverse benchmarks show that LaV-CoT outperforms comparable open-source models and rivals larger or proprietary ones. Furthermore, deployment on an Intelligent Document Processing platform demonstrates practical gains in accuracy, user satisfaction, and efficiency.


While promising, our approach has several limitations. Coverage of low-resource languages remains constrained by the scarcity of high-quality multilingual data. Additionally, the current framework focuses on slow, step-by-step reasoning and does not yet support fast or hybrid reasoning. In future work, we aim to extend LaV-CoT to broader low-resource and domain-specific scenarios, incorporate fast and hybrid reasoning mechanisms, and enhance multi-aspect reward modeling to build more robust and inclusive multimodal reasoning systems.

\section{Ethical Use of Data}

All examples in this work, including those in the Appendix, are drawn from open-source datasets or anonymized real-world data, ensuring no private information is disclosed.

\section{Acknowledgments}
This work was supported by the Ant Group Postdoctoral Programme. This work was also supported by the Japan Science and Technology Agency (JST) and the Agency for Science, Technology and Research (A*STAR) under the Japan-Singapore Joint Call (Project No. R24I6IR133).


\bibliographystyle{ACM-Reference-Format}
\bibliography{sample-base}

\newpage
\appendix

\section{Prompt Design}
\label{app:prompt_design}
\subsection{Generator Prompt}
\begin{figure}[H]
  \centering
  \includegraphics[width=\linewidth]{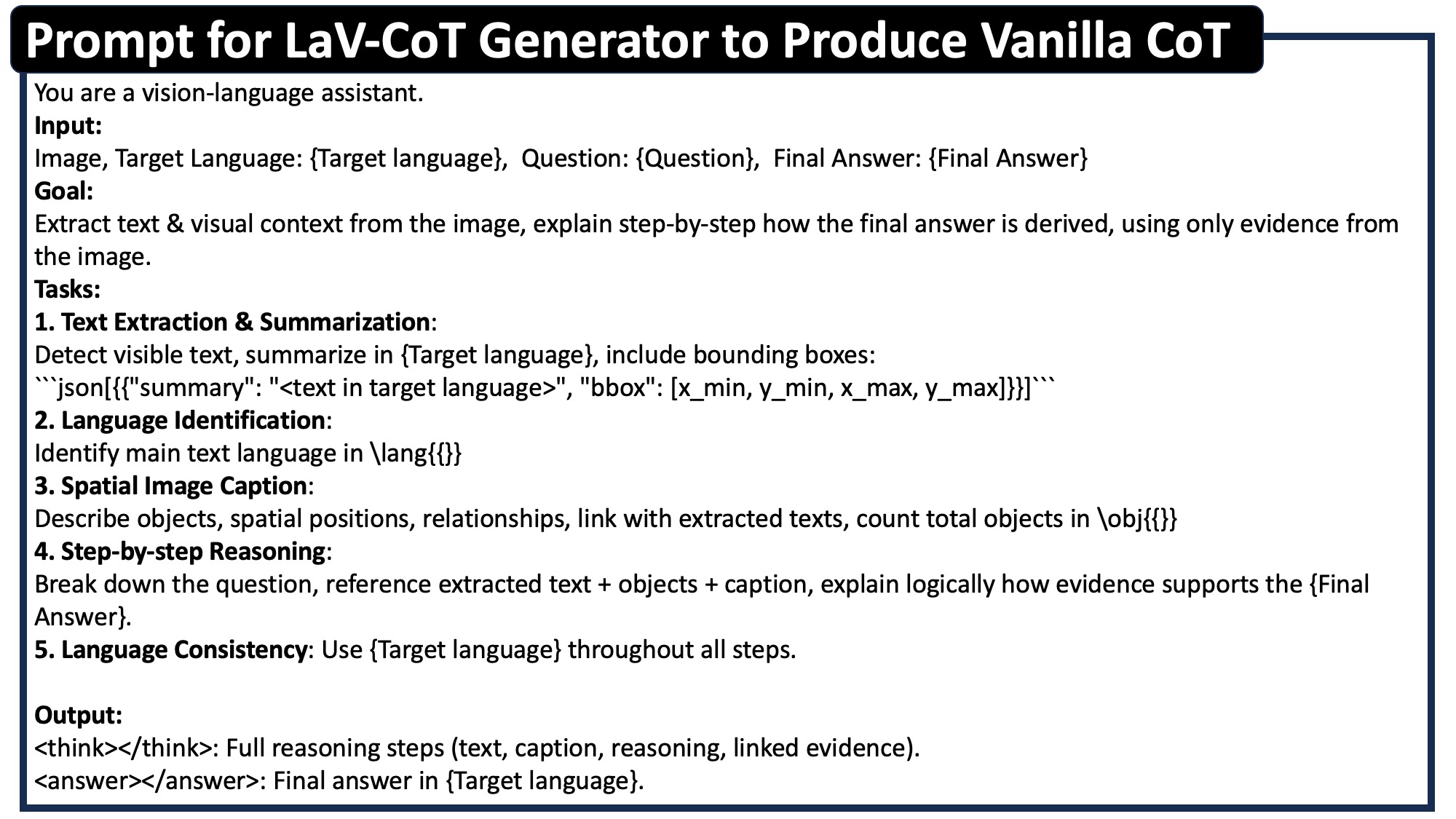}
  \caption{Instructions for generating Vanilla CoT.}
  \label{prompt:vanilla cot}
  \Description{Vanilla CoT Prompt.}
\end{figure}
\begin{figure}[H]
  \centering
  \includegraphics[width=\linewidth]{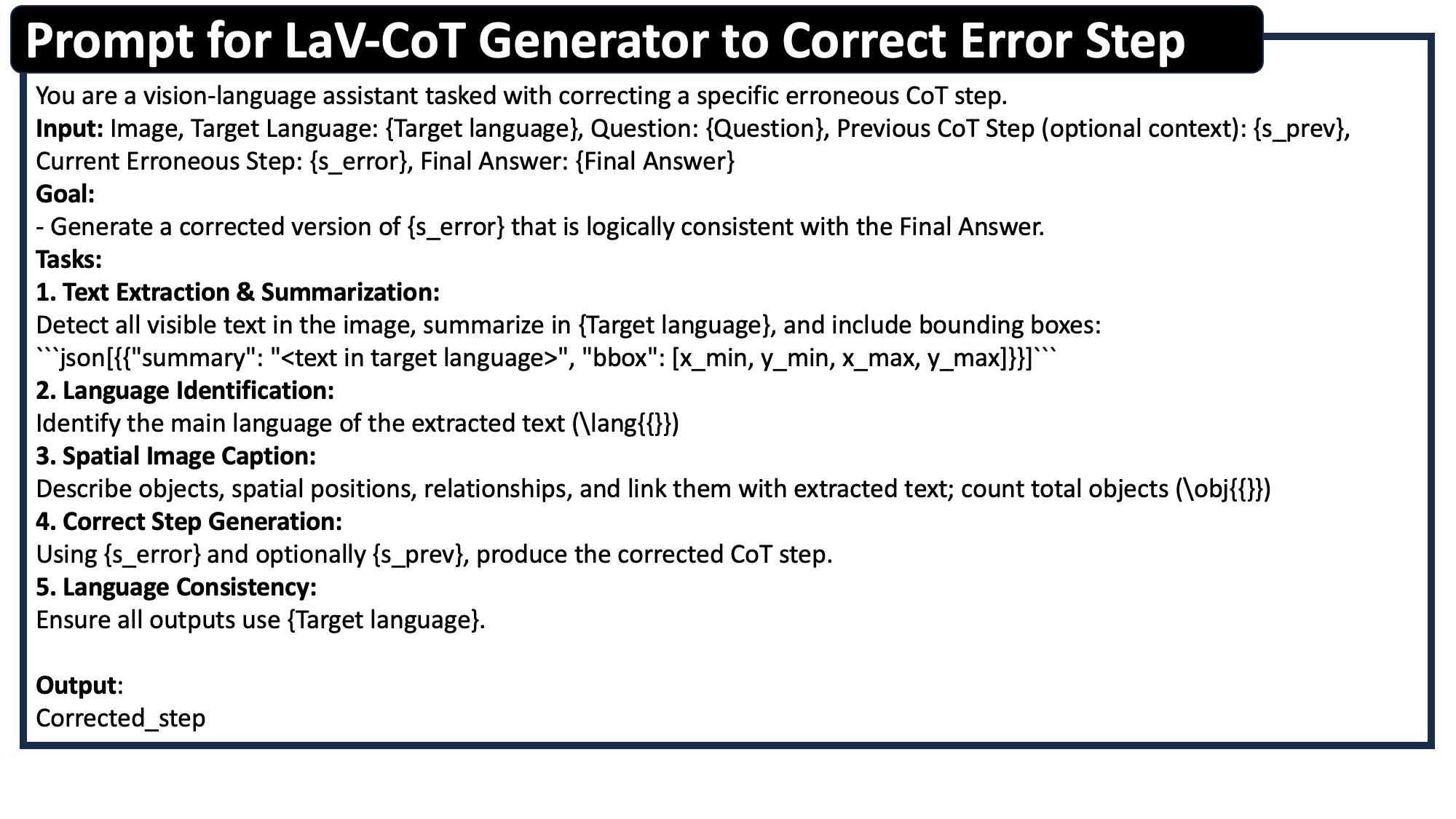}
  \caption{Instructions for correcting error steps.}
  \label{prompt:error step}
  \Description{Correcting error CoT Prompt.}
\end{figure}
\subsection{Evaluator Prompt}
\begin{figure}[H]
  \centering
  \includegraphics[width=\linewidth]{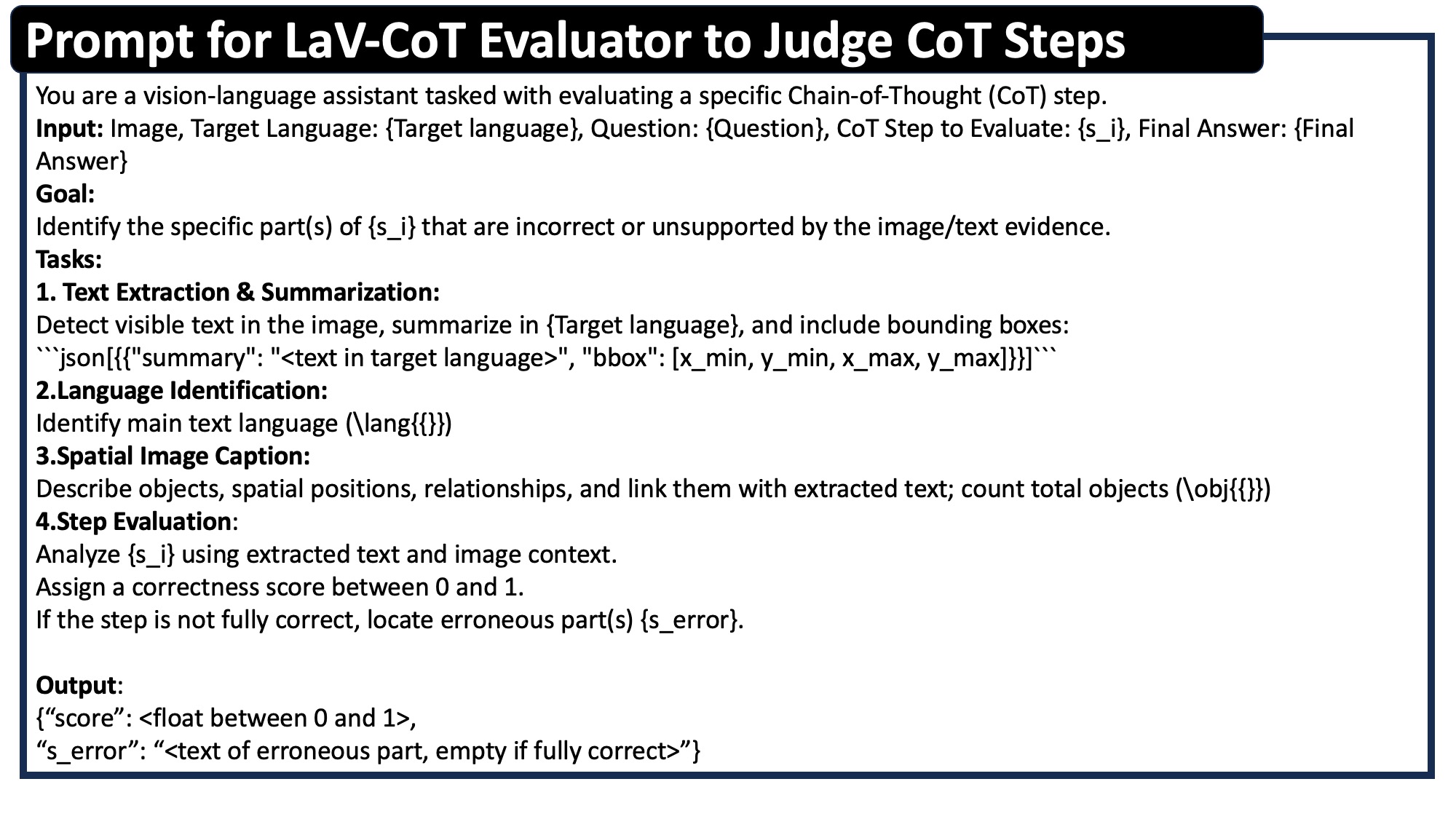}
  \caption{Instructions for evaluating CoT steps.}
  \label{prompt:evaluate step}
  \Description{Evaluating CoT Prompt.}
\end{figure}

\subsection{Inference Prompt}
\begin{figure}[H]
  \centering
  \includegraphics[width=\linewidth]{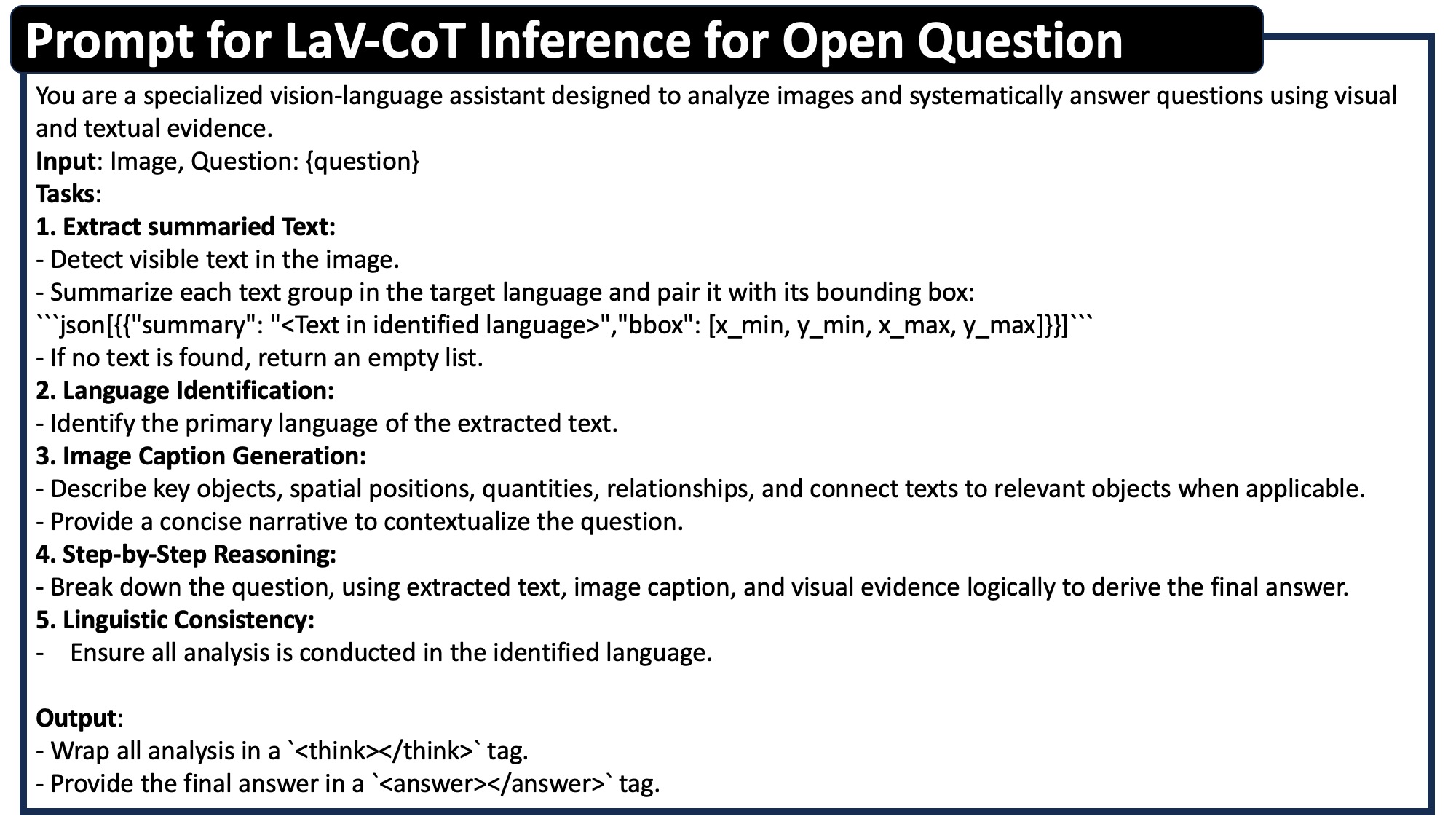}
  \caption{Instructions for LaV-CoT inference on open-ended questions.}
  \label{prompt:inference prompt open}
  \Description{Instructions for LaV-CoT inference on open-ended questions.}
\end{figure}

\begin{figure}[H]
  \centering
  \includegraphics[width=\linewidth]{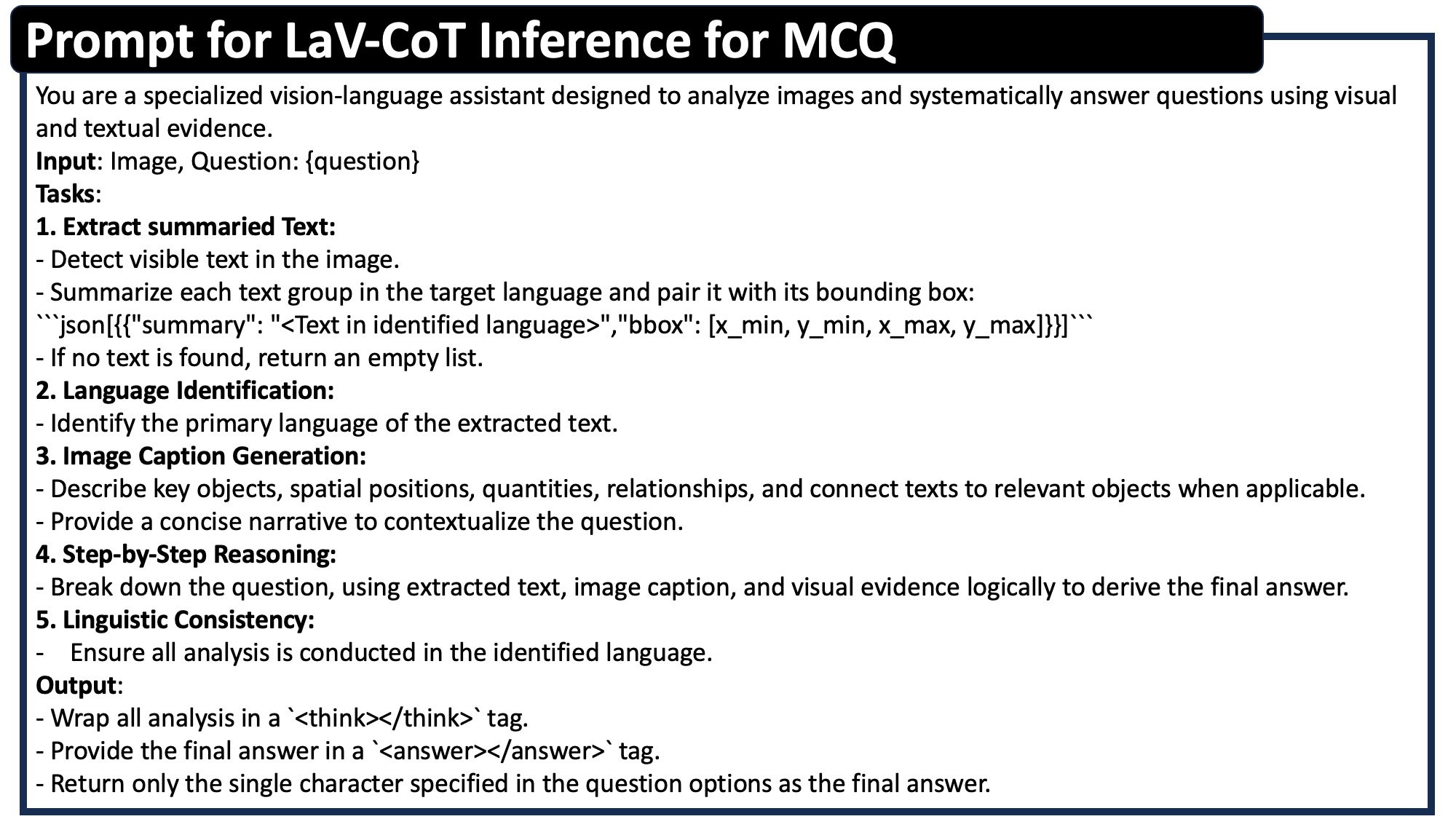}
  \caption{Instructions for LaV-CoT inference on MCQ.}
  \label{prompt:inference prompt mcq}
  \Description{Instructions for LaV-CoT inference on MCQ.}
\end{figure}

\section{Additional Performance Comparison}
\label{app:add_performance}
\begin{table}[h]
  \centering
  \label{tab:add_results}
  \begin{tabular}{l ccc}
    \toprule
    Model & DocVQA & ChartQA & OCRBenchv2(en/cn) \\
    \midrule
    Qwen2.5-VL-3B       & 93.9 & 83.9 & 52.1/54.3 \\
    Qwen2.5-VL-7B       & 94.1 & 84.2 & 53.2/56.1 \\
    GPT-4o-2024-05-13         &  94.4 & 85.1 & 55.8/57.6 \\
    Gemini-2.5-flash    & 94.9 & 86.0 & 56.0/57.9 \\
    \midrule
    LaV-CoT-3B(SFT)     & 94.5 & 85.4 & 55.2/57.6 \\
    LaV-CoT-3B(SFT+GRPO) & 95.0 & 86.3 & 56.4/58.2 \\
    \bottomrule
  \end{tabular}
  \caption{Comparison of different models on standard document understanding benchmarks.}
\end{table}
\section{More Real-world Cases}
Please see the next page for more cases.
\begin{figure*}[h] 
  \centering
  \includegraphics[width=0.85\textwidth]{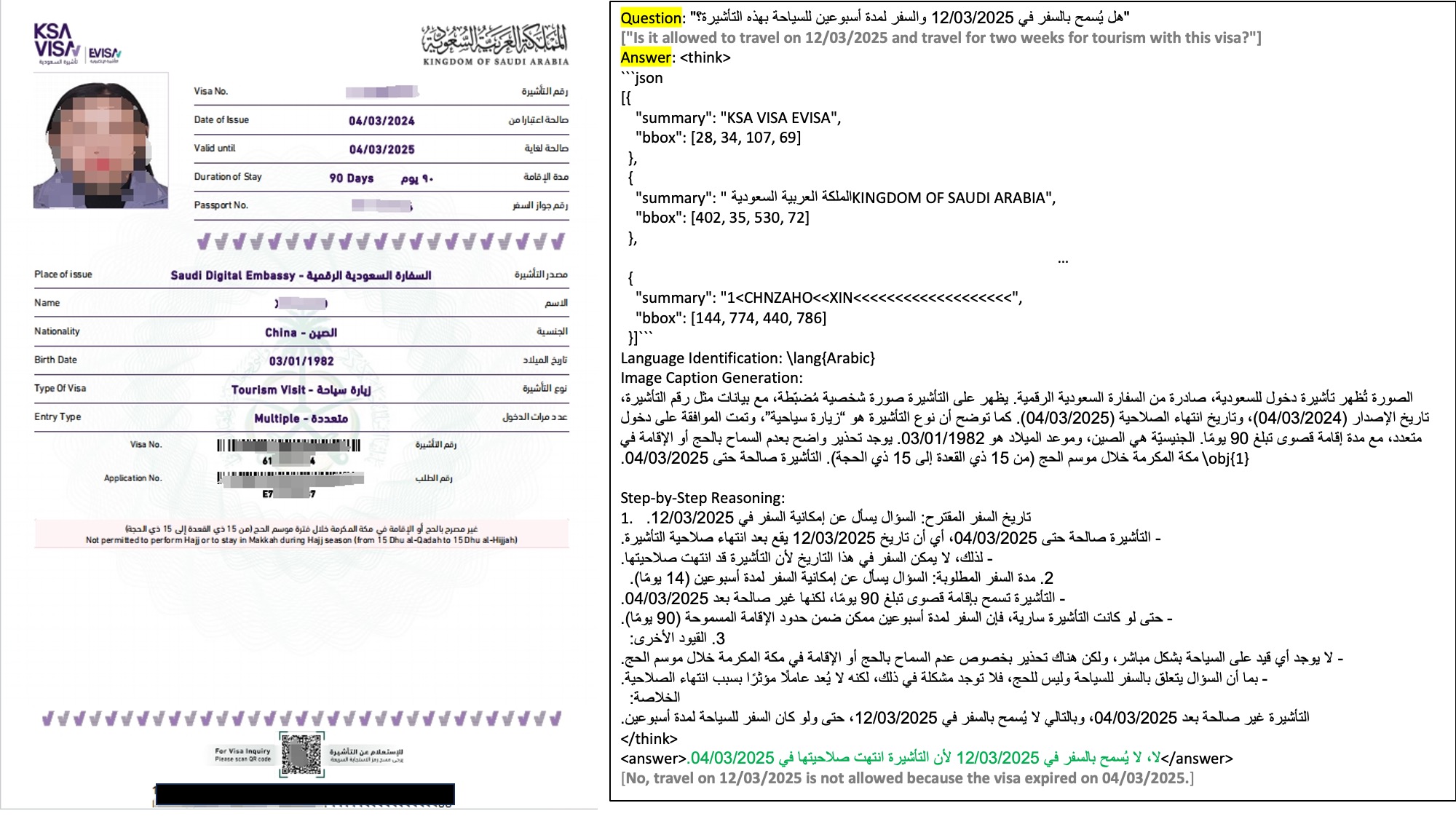}
  \caption{Arabic Visa demo case.}
  \label{prompt:arabic_visa}
  \Description{Arabic visa demo case.}
\end{figure*}

\begin{figure*}[h]
  \centering
  \includegraphics[width=0.85\textwidth]{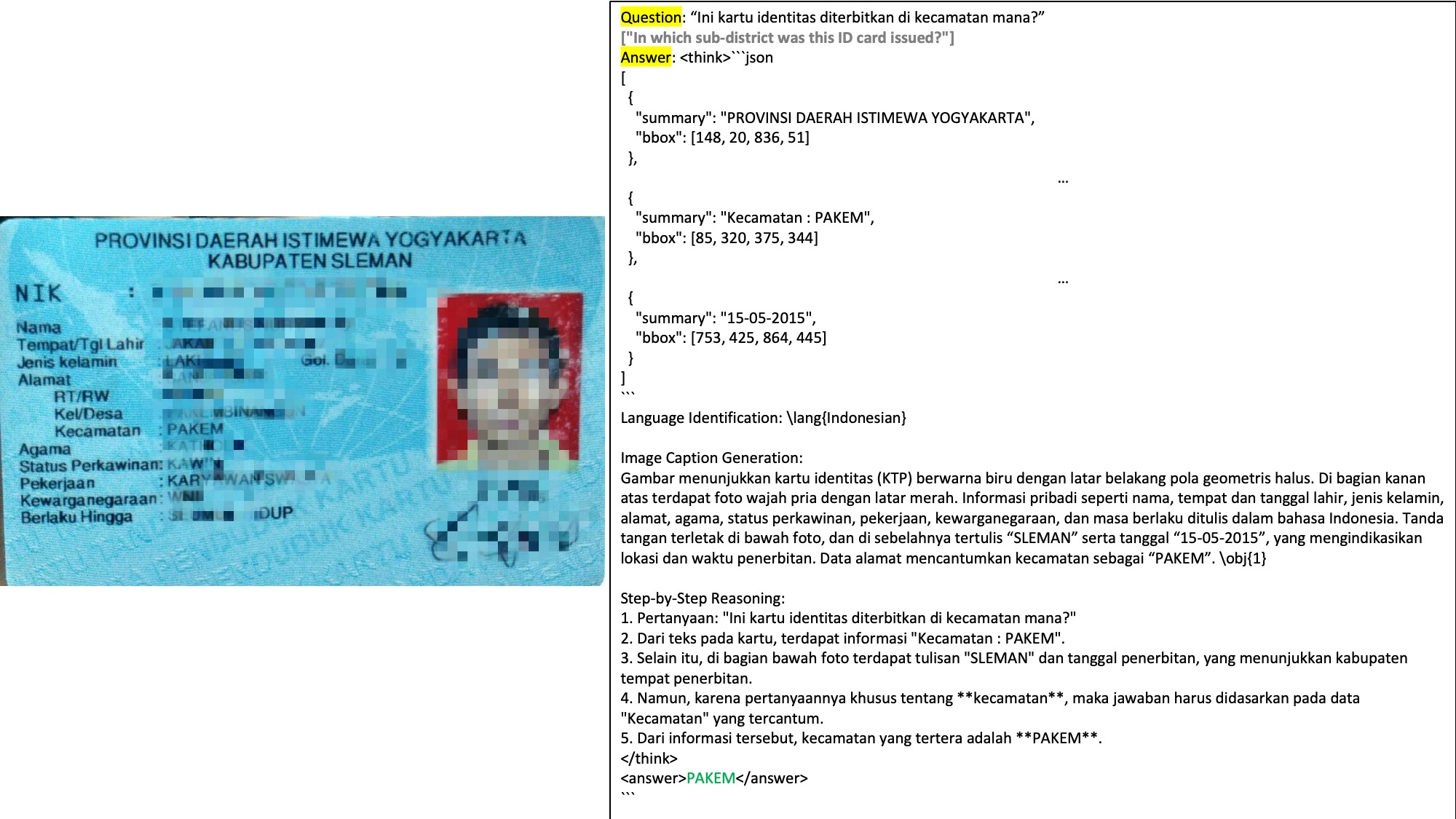}
  \caption{Indonesian ID demo case.}
  \label{prompt:indo_id}
  \Description{Indonesian ID demo case.}
\end{figure*}
\end{document}